\documentclass[letterpaper, 10 pt, conference]{ieeeconf}  

\IEEEoverridecommandlockouts                              

\usepackage{graphicx}
\usepackage{subcaption}
\usepackage{amsmath}
\usepackage{amssymb}
\usepackage{booktabs}
\usepackage{url}
\usepackage[hidelinks]{hyperref}

\title{\LARGE \bf
Doppler-aided Position-Velocity Tracking \\ of Hidden Objects using mmWave Radar
}

\author{Yerin Cheon$^{1*}$, Kush Garg$^{2*}$, and Akshat Dave$^{1}$%
\thanks{$^{1}$Stony Brook University,
100 Nicolls Rd, Stony Brook, NY 11794, United States.
{\tt\small \{ycheon, dave\}@cs.stonybrook.edu}}%
\thanks{$^{2}$Delhi Technological University, 110042, New Delhi, India.
{\tt\small kushgarg\_23mc079@dtu.ac.in}}%
\thanks{*These authors contributed equally.}%
}

\usepackage{amsmath}

\IEEEaftertitletext{%
  \vspace{-0.5\baselineskip}%
  \begin{minipage}{\textwidth}%
    \centering%
    \includegraphics[width=\textwidth]{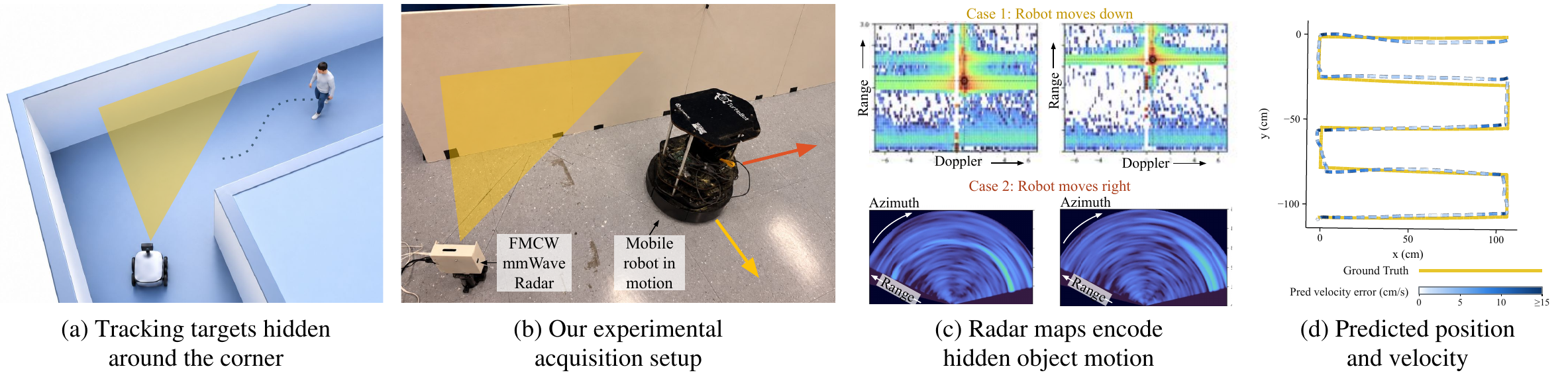}%
    \captionof{figure}{\textbf{Doppler-aided Radar NLOS Tracking.} Tracking objects hidden around corners enables safe robot operation (a). We consider an FMCW radar that faces a wall, and its indirect reflections bounce back from a moving hidden mobile robot (b). The measured Range-Azimuth-Doppler (RAD) maps of the radar encode the motion of the hidden object (c). We capture a dataset of RAD maps synchronized with the object's position and velocity. We develop a data-driven framework that simultaneously predicts position and velocity along the trajectory at high accuracy on experimental captures (d).}
    \label{intro}%
  \end{minipage}%
  \vspace{0.5\baselineskip}%
}

\begin{document}

\maketitle
\thispagestyle{empty}
\pagestyle{empty}


\begin{abstract}
Safe indoor operation requires autonomous tracking of objects that are not directly visible to the robot. Optical non-line-of-sight (NLoS) methods depend on ambient lighting and scene geometry or on weak multi-bounce returns that need long acquisitions. Millimeter-wave (mmWave) radar offers an attractive alternative: walls act as reflectors for mmWave signals, and frequency-modulated continuous-wave (FMCW) radars enable fast, practical measurement of range, angle, and velocity. However, the low resolution of these radars leads to spatially ambiguous observations. The Doppler information, although particularly sensitive to target motion, remains underexplored in indoor NLoS localization: existing methods emphasize spatial representations and either collapse or discard the Doppler dimension before learning. To leverage the complete Range--Azimuth--Doppler (RAD) information, we acquire the first indoor NLoS radar dataset that pairs sequential RAD measurements with synchronized position and velocity ground truth under controlled variations in motion: 576 sequences totaling 720,856 frames across 12 trajectories of a mobile robot moving along varying speeds, starting positions, and material configurations. We then formulate NLoS tracking as sequential position-velocity estimation rather than frame-wise localization: a shared 3-D encoder processes the full RAD representation of consecutive frames, separate recurrent branches estimate position and velocity, and post-inference refinement combines the two into a motion-consistent trajectory. Our method achieves 8.95\,cm position and 1.75\,cm/s velocity errors on held-out trajectories and generalizes to unseen scenarios involving a moving human target. Together, our dataset and tracker establish a benchmark for tracking hidden moving targets with a low-cost radar. Project page: \href{https://ycheon1101.github.io/yerincheon/radar_nlos/}{\nolinkurl{ycheon1101.github.io/yerincheon/radar_nlos/}}

\end{abstract}
\section{Introduction}

Safe indoor operation requires autonomous tracking of objects that are not directly visible to the robot. For instance, in warehouses, where humans and robots operate in close proximity, corners and corridors create blind spots, increasing the risk of collisions. Non-line-of-sight (NLoS) tracking aims to leverage indirect reflections of light to estimate the position of hidden targets~\cite{somasundaram2026imaging,young2025enhancing}.
Optical NLoS methods, whether
  camera-based~\cite{yedidia2019using,czajkowski2024two,chen2019steady} or
  LiDAR-based~\cite{young2025enhancing,somasundaram2026imaging,garg2025superex}, depend on ambient lighting and
  scene geometry~\cite{wang2021accurate} or on weak multi-bounce returns that need long
  acquisitions, which limits range and real-time tracking.

Millimeter-wave (mmWave) radar provides an attractive alternative. RF signals are largely insensitive to illumination conditions, and walls can act as reflectors for mmWave signals, creating strong indirect paths toward regions outside direct visibility. In particular, Frequency-modulated continuous-wave (FMCW) mmWave radars have recently become practical through mass production and provide fast, continuous measurements of the range, angle of arrival, and velocity of objects. 

However, translating FMCW radar measurements into accurate NLoS tracking remains challenging. Indirect propagation produces spatially ambiguous radar observations, so the relationship between the measured radar response and the hidden target position difficult to model explicitly. Moreover, the limited spatial resolution of low-cost FMCW radar makes fine-grained target localization challenging. To address this spatial ambiguity, we adopt a data-driven approach that directly learns the relationship between NLoS radar measurements and target position.

For moving targets, consecutive radar observations provide temporal diversity that is unavailable from a single frame. As the target moves, changes in its position and Doppler response generate complementary observations over time, providing temporal information for continuous NLoS tracking. Among these measurements, Doppler is particularly sensitive to target motion, yet it remains relatively underexplored in indoor radar NLoS localization. This is partly because NLoS propagation distorts the relationship between measured Doppler and the target's physical velocity, and because Doppler responses can weaken when the target slows down, changes direction, or temporarily stops. Existing indoor methods have therefore largely emphasized spatial representations such as range--angle or angle--delay maps~\cite{yue2022cornerradar,xu2024tracking}. The Doppler dimension is collapsed into a per-point attribute~\cite{scheiner2019seeing} or discarded before learning~\cite{xu2024tracking}. 

To leverage the complete Range--Azimuth--Doppler (RAD) information for NLoS tracking, we acquire real measurements that pair sequential RAD observations with synchronized position and velocity ground truth across diverse motion. Existing real-world NLoS radar datasets~\cite{xu2024tracking,10842463,scheiner2019seeing,yue2022cornerradar,lai2026non,du2026x} are designed for specific environments or motion scenarios, and do not provide full RAD--position--velocity information under controlled variation in target motion. We place a stationary mmWave radar in front of a wall and move a mobile robot along varying trajectories, speeds, and starting positions, with varying material properties, yielding 576 sequences comprising 720,856 radar frames. To our knowledge, this is the first indoor NLoS radar dataset to provide sequential RAD measurements with both position and velocity labels under controlled motion variation.

Along with this dataset, we formulate NLoS tracking as a sequential position-velocity estimation problem rather than independent frame-wise localization. Our model preserves the full RAD representation and processes a short sequence of consecutive frames. A shared 3-D encoder extracts features from each frame, followed by separate recurrent branches for position and velocity. We found that coupling the position state to velocity-based updates within the network caused localization errors to accumulate, even when the predicted velocity was accurate. We therefore combine position and velocity estimates after inference, refining the trajectory to agree with both predicted locations and inter-frame motion. Low predicted speed suppresses unnecessary movement, reducing drift and stationary jitter.

Our main contributions are as follows:

\begin{itemize}

\item A data-driven indoor \textbf{NLoS tracking framework} that jointly estimates position and velocity from sequences of full RAD measurements, through independent estimation followed by a motion-consistent refinement stage. Our method achieves mean position and velocity errors of $8.95$\,cm and $1.75$\,cm/s, respectively, on held-out validation recordings, and also generalizes to unseen trajectories of a moving person.

\item A controlled real-world \textbf{NLoS radar dataset} of 576 diverse sequences spanning varied trajectories, speeds, starting positions, and material configurations. Each sequence provides RAD measurements together with synchronized position and velocity ground truth, enabling systematic evaluation of generalization to held-out motion and target conditions.
\end{itemize}
\section{Related Work}
\subsection{Optical Non-Line-of-Sight Sensing and Localization}
Optical NLoS methods demonstrated hidden-scene reconstruction by exploiting time-resolved measurements of multiply reflected light~\cite{kirmani2009looking,velten2012recovering,gupta2012reconstruction}, however, typically required dense spatial measurements and specialized time-resolved hardware. Subsequent work improved the practicality and computational efficiency of active NLoS imaging through SPAD-based acquisition, confocal scanning, and wave-based reconstruction formulations~\cite{buttafava2015non,o2018confocal,lindell2019wave,liu2019non}. Beyond static reconstruction, Gariepy et al.~\cite{gariepy2016detection} directly localized and tracked moving hidden targets, while later work extended optical NLoS tracking to longer-range scenarios~\cite{chan2017non}. To reduce active-hardware requirements, passive approaches exploited radiance variations, occlusion cues, and temporal correlations using conventional cameras or ambient light~\cite{bouman2017turning,saunders2019computational,boger2019passive, czajkowski2024two}, although their performance remained sensitive to illumination, geometry, and weak indirect signals. More recent systems improved dynamic reconstruction using low-latency SPAD sensing and consumer-grade LiDAR with multi-frame integration~\cite{nam2021low,ye2024real,somasundaram2026imaging}. Despite these advances, optical approaches remain constrained by weak multi-bounce returns and sensing requirements, motivating the use of RF signals for robust NLoS localization and tracking.

\subsection{RF-Based NLoS Sensing}

RF-based NLoS sensing has traditionally exploited reflected and multipath propagation to detect and localize targets beyond direct visibility~\cite{sume2011radar,zetik_nlos,thai_nlos,zhao2018millimeter}. Model-based approaches infer hidden-target positions from known scene geometry, phase relationships, virtual targets, or explicit multipath models and propagation-path associations~\cite{zetik_nlos,thai_nlos,guo2020behind,yang2020nlos,li2021multiple,tang_nlos}. Although effective in structured settings, these methods rely on accurate geometric assumptions and propagation-path association, which become increasingly fragile in complex indoor environments. Efforts to reduce this dependence include jointly estimating scene structure and hidden targets~\cite{chen_nlos}. More recently, CornerRadar~\cite{yue2022cornerradar} showed that ray-tracing-based localization can suffer from mirroring and splitting effects in realistic indoor scenes, motivating learning-based RF localization that models complex NLoS propagation directly from measurements. In contrast to approaches that explicitly model or associate individual propagation paths, our method learns directly from sequential full Range--Azimuth--Doppler measurements for continuous indoor NLoS localization and motion estimation.

\subsection{Learning-Based RF NLoS Tracking}

Learning-based approaches reduce reliance on explicit propagation models by directly learning the mapping from indirect RF measurements to hidden-target locations. Recent work extends learned radar NLoS sensing to 3D scene reconstruction with cascaded mmWave radar~\cite{lai2026non} and X-band arrays~\cite{du2026x}, but targets static scene structure rather than the state of a moving target. CornerRadar~\cite{yue2022cornerradar} combines RF measurements with geometry-derived hint maps to address mirroring and splitting effects, while Xu et al.~\cite{xu2024tracking} estimate target positions from angle--delay or range--angle representations using a CNN followed by Kalman filtering, and report degraded localization performance when range--Doppler measurements are directly incorporated. These indoor approaches primarily focus on position estimation from spatial radar representations rather than jointly learning target motion, highlighting the difficulty of exploiting Doppler under NLoS propagation. Although Doppler-aware tracking has been explored in automotive NLoS settings~\cite{scheiner2019seeing}, such methods typically operate on sparse radar detections in road-specific scenarios. In contrast, we learn directly from sequential full Range--Azimuth--Doppler (RAD) representations, jointly exploiting spatial cues, Doppler information, and temporal diversity to estimate both target position and velocity for continuous indoor three-bounce NLoS tracking.

\begin{figure}[t]
\centering
\begin{subfigure}{\linewidth}
    \centering
    \includegraphics[width=\linewidth]{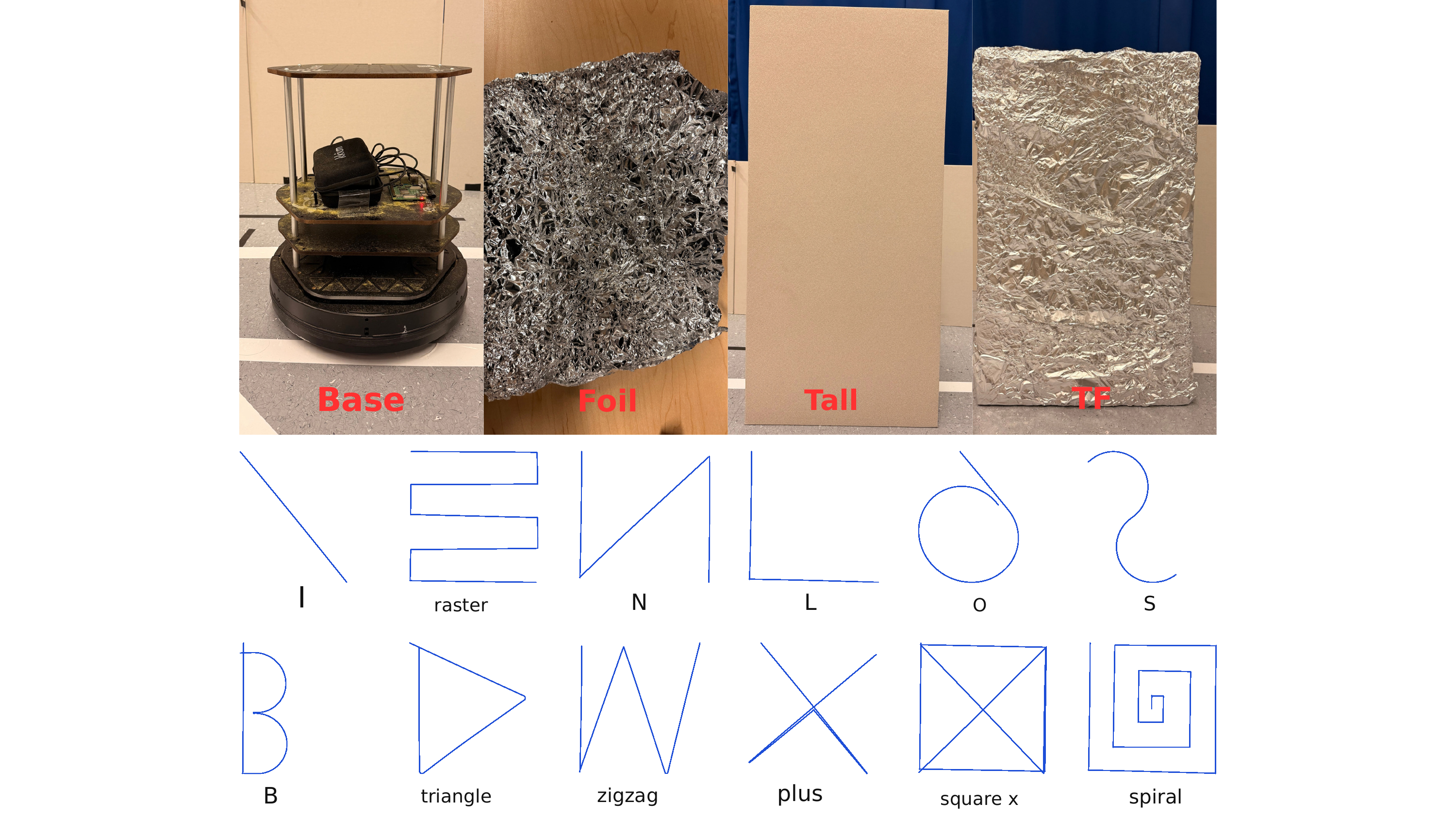}
    \label{materials_trajs}
\end{subfigure}

\caption{\textbf{Materials and trajectories in our dataset acquisition}. Our dataset includes four target configurations and 12 trajectory patterns.
}
\label{fig:setup}
\end{figure}

\section{Data Collection and Calibration}
\label{sec:data_collection}

\subsection{Experimental Setup and Radar Configuration}
We construct an mmWave radar dataset for indoor three-bounce NLoS localization and tracking of moving targets. The data collection system consists of a TI AWR1843 FMCW radar, a DCA1000 raw ADC capture board, a 2.7-cm-thick drywall panel, and a TurtleBot/Kobuki mobile robot used as the moving target. The target is placed outside the radar's direct field of view (FoV), and the radar observes it through a three-bounce wall--target--wall propagation path.

Each trial contains a sequential Range--Azimuth--Doppler (RAD) representation generated from the raw ADC measurements, together with synchronized ground-truth position $(x,y)$ and velocity $(v_x,v_y)$ obtained from Kobuki wheel odometry. All ground-truth trajectories are transformed into a common desk-fixed coordinate frame to provide a consistent physical reference across different initial headings, with trajectory-specific preparation headings used for alignment.

The radar operates in the 76 GHz band, and each radar frame consists of 128 chirps and four receiver channels. The theoretical range resolution is approximately 8.33 cm, and radar measurements are acquired at 20 Hz. Ground-truth position and velocity are temporally aligned with the corresponding radar timestamps.

\subsection{Trajectory and Data Collection Protocol} 
To evaluate generalization across diverse motion and target conditions, our dataset systematically varies trajectory, velocity, starting position, and target configuration. We collect 12 trajectory patterns: I-shaped, raster scan, N, L, O, S, B, triangle, zigzag, plus, square-X, and square spiral. Each trajectory is recorded at three nominal velocities of 0.25, 0.75, and 1.00~m/s and from four starting positions to expose the model to different motion directions and spatial configurations. Most trajectories are executed within an approximately $1.5,\mathrm{m}\times1.5,\mathrm{m}$ workspace, with the robot stopping to rotate whenever a change in heading is required. We further vary the target configuration using four materials: the standard TurtleBot, a foam-board target, a target covered with crumpled aluminum foil, and a foam-board target wrapped with crumpled aluminum foil. The two foam-board configurations use different heights to introduce additional variation in target elevation and radar reflectivity. The trajectory patterns and target configurations are illustrated in Fig.~\ref{fig:setup}. Combining 12 trajectories, three velocities, four target configurations, and four starting positions results in a total of 576 recorded sequences. Depending on the trajectory length and motion pattern, each sequence contains approximately 350 to 4,200 radar frames together with synchronized position and velocity ground truth.

\subsection{Calibration}
We perform three calibration and preprocessing steps: spatial calibration, temporal synchronization, and empty-scene background suppression. Spatial calibration is used to obtain consistent ground-truth labels across all trials by transforming the robot odometry into a fixed desk coordinate frame. Temporal synchronization aligns the Raspberry Pi used for robot motion logging with the local device used for radar capture. Both systems record timestamps in Unix epoch time, which provides a common temporal reference. To obtain a more robust alignment between robot motion onset and radar acquisition, we further refine the synchronization through manual post-processing after data collection.

Static background suppression is commonly used in radar processing to remove persistent environmental reflections and retain more informative target-dependent measurements. In our setting, however, Doppler information is particularly important. Aggressive clutter removal may suppress useful responses in low-Doppler regions when the target moves slowly, rotates, or temporarily stops, whereas insufficient suppression can leave strong background responses that may lead to erroneous localization. We therefore use a statistical thresholding strategy based on empty-scene measurements rather than directly removing low-Doppler bins. The details of this procedure are described in Sec.~\ref{background_noise}.

\begin{figure*}[t]
    \centering
    \includegraphics[
        width=\textwidth,
        clip
    ]{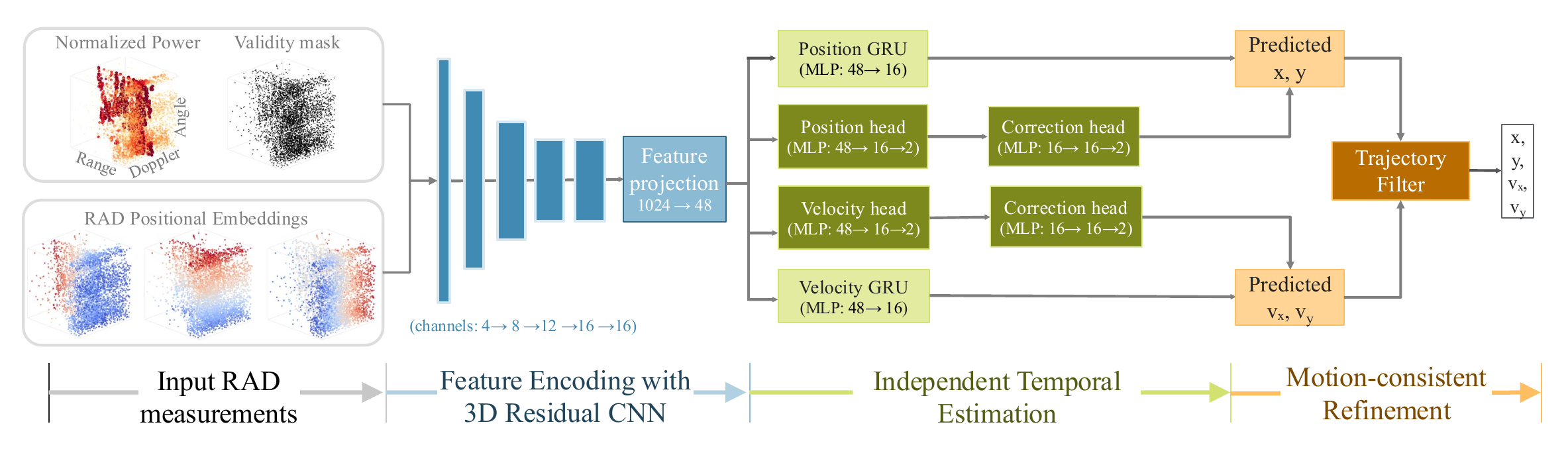}
    \caption{\textbf{Our NLoS position-velocity tracking pipeline.}  Input RAD frame comprises 5 channels for each time step: normalized power, validity mask, and positional embeddings for RAD dimensions. A shared 3D residual CNN encodes the RAD frame into a 48-D feature. Independent position and velocity branches add a gated recurrent unit (GRU)- based temporal correction. The GRU hidden state is updated through eight consecutive RAD frames. The resulting position and velocity estimates are then refined for motion consistency.}
    \label{fig:architecture}
\end{figure*}
\section{Method}
\label{sec:method}

Our method operates on sequential Range--Azimuth--Doppler (RAD) representations constructed from raw FMCW radar measurements. The processing pipeline converts the raw ADC signals into range, azimuth, and Doppler dimensions while suppressing static background responses using the empty-scene calibration described in Sec.~\ref{background_noise}. We preserve the full RAD structure rather than reducing the measurements to spatial maps or sparse detections, allowing the subsequent temporal model to jointly exploit spatial and motion information. The resulting RAD sequences are then used to jointly estimate the position and velocity of the hidden target.

\subsection{RAD Construction}

Raw ADC measurements are converted into Range--Azimuth--Doppler (RAD) representations using standard FMCW processing. A Hann-windowed FFT along the fast-time dimension produces the range response, followed by a Doppler FFT across chirps. Angular information is then obtained from the four receiver channels using a zero-padded 64-point spatial FFT. The resulting power tensor has dimensions $61\times64\times128$, corresponding to range, azimuth, and Doppler, respectively.

\subsection{Background Suppression}
\label{background_noise}
Static wall reflections and environmental clutter can dominate the weak NLoS target response. We therefore construct an empty-scene background model and estimate a separate threshold for each range--Doppler bin to account for spatially varying background responses. The threshold is defined as
\begin{equation}
T(r,d)
=
\mu_{\mathrm{empty}}(r,d)
+
3\sigma_{\mathrm{empty}}(r,d).
\label{eq:background_threshold}
\end{equation}
RAD responses below this threshold are removed. Unlike hard suppression of zero- or low-Doppler bins, this statistical thresholding can retain low-Doppler target responses when they exceed the corresponding empty-scene background level. The remaining valid RAD responses are stored sparsely without additional range, azimuth, or Doppler downsampling.

\subsection{Radar Feature Encoding}
We estimate the hidden target state  $\mathbf{s}_t = [x_t, y_t, v_{x,t}, v_{y,t}]^\top$ in the fixed desk-coordinate system. Each radar frame is represented by normalized log-power and a binary validity mask, giving an input of size [$2\times61\times64\times128$]. Power is clipped and normalized using fixed training-set statistics. The mask distinguishes thresholded cells from valid measurements. Three fixed RAD positional-embedding channels containing normalized range, azimuth, and Doppler coordinates are concatenated with the power and validity-mask channels, providing the network with the location of each response within the RAD volume.
As shown in Fig.~\ref{fig:architecture}, a shared 3-D residual CNN processes each frame independently. An initial convolution reduces the Doppler dimension, followed by four residual downsampling blocks with $8$, $12$, $16$, and $16$ output channels. Group normalization and SiLU activations are used throughout. The resulting $16\times4\times4\times4$ feature volume is flattened and projected to a vector $\mathbf{z}_t\in\mathbb{R}^{48}$. Flattening retains coarse spatial arrangement for localization in the fixed experimental setup. Two frame-level regression heads map $\mathbf{z}_t$ to initial position and velocity estimates, while the same embedding is passed to the temporal module.

\subsection{Independent Temporal Position and Velocity Estimation}
Temporal observations provide additional evidence when individual radar returns are ambiguous, as exploited in prior NLoS radar systems~\cite{scheiner2019seeing,yue2022cornerradar}.

We process eight consecutive frames using two independent, single-layer gated recurrent units (GRUs), each with a 16-dimensional hidden state. Both receive the shared feature sequence, while their recurrent states and prediction heads are separate. This allows position and velocity to use different temporal cues without forcing absolute position to follow velocity integration.
Each branch combines a single-frame estimate with a bounded temporal correction. For task $k\in\{p,v\}$,
\begin{equation}
\begin{aligned}
\mathbf{h}_t^k
&= \operatorname{GRU}_k
   \left(\mathbf{z}_t,\mathbf{h}_{t-1}^k\right),\\
\hat{\mathbf{u}}_t^k
&= f_k(\mathbf{z}_t)
 + \tanh\!\left(g_k(\mathbf{h}_t^k)\right).
\end{aligned}
\label{eq:temporal_residual}
\end{equation}
Here, $\hat{\mathbf{u}}_t^k$ denotes a prediction in standardized target coordinates. The single-frame heads $f_k$ have dimensions $48\rightarrow16\rightarrow2$, and the correction heads $g_k$ have dimensions $16\rightarrow16\rightarrow2$. Corrections are bounded to one training-set standard deviation per output component. Each correction head receives only its corresponding GRU state. Recurrent states are reset between non-overlapping eight-frame clips, and a prediction is produced at every valid timestep. The complete network contains $95{,}164$ trainable parameters.
Position and velocity targets are standardized independently using training-set means and standard deviations. We train all neural components jointly with loss:
\begin{equation}
\mathcal{L}
=
10\,\operatorname{SmoothL1}
\!\left(\hat{\mathbf{u}}^{p},\mathbf{u}^{p}\right)
+
10\,\operatorname{SmoothL1}
\!\left(\hat{\mathbf{u}}^{v},\mathbf{u}^{v}\right).
\label{eq:state_loss}
\end{equation}
averaged over valid timesteps and output components, with transition parameter $\beta=1$. Predictions are subsequently converted to physical units.

\subsection{Motion-Consistent Trajectory Refinement}
In preliminary experiments, we used predicted velocity to constrain inter-frame position changes through a displacement-consistency loss. Although this improved trajectory smoothness, velocity errors could accumulate over time and lead to persistent position offsets, particularly during stationary intervals. Such offsets cannot be corrected from velocity alone when the absolute position estimate is inaccurate. We therefore retain separate neural estimates of absolute position and velocity and apply motion consistency through trajectory-optimization.
Let $\hat{\mathbf{p}}_t$ and $\hat{\mathbf{v}}_t$ denote the network predictions in physical units. A five-frame median filter is applied to predicted speed; frames below $1 \mathrm{cm/s}$ are classified as stationary. Setting predicted velocity to zero at these frames gives $\bar{\mathbf{v}}_t$. The expected displacement is $\mathbf{d}t=\tfrac{\Delta t}{2}(\bar{\mathbf{v}}_{t-1}+\bar{\mathbf{v}}_t)$, where $\Delta t=0.05,\mathrm{s}$. We estimate the refined positions by approximately minimizing
\begin{equation}
\begin{aligned}
\mathbf{q}_{1:T}^{\star}
=\arg\min_{\mathbf{q}_{1:T}}\;&
\sum_{t=1}^{T} a_t\,
\rho_{\delta}\!\left(
\left\|\mathbf{q}_t-\hat{\mathbf{p}}_t\right\|_2
\right)\\
&+\frac{1}{2}\sum_{t=2}^{T}\lambda_t
\left\|\Delta\mathbf{q}_t-\mathbf{d}_t\right\|_2^2 .
\end{aligned}
\label{eq:trajectory_refinement}
\end{equation}
Here, $\Delta\mathbf{q}_t=\mathbf{q}_t-\mathbf{q}_{t-1}$ and $\mathbf{d}_t$ is the displacement obtained from the predicted velocity.
The Huber penalty $\rho\delta$ limits the influence of inconsistent position predictions. We set $a_t=10^{-3}$ at stationary frames and $a_t=1$ otherwise. The motion weight is $\lambda_t=10^6$ when both adjacent frames are stationary and $100$ otherwise, with $\delta=5,\mathrm{cm}$. Five iteratively reweighted least-squares solves yield the refined trajectory.
For the reported evaluation, final velocity is obtained by centered finite differences of $\mathbf{q}_t^\star$, using one-sided differences at the endpoints and zero velocity at detected stops. The same refinement objective can be applied continuously in a fixed-lag sliding window as new radar observations arrive, with backward finite differences used for causal velocity estimation.
\begin{figure}[t]
    \centering
    \includegraphics[
        width=\columnwidth
    ]{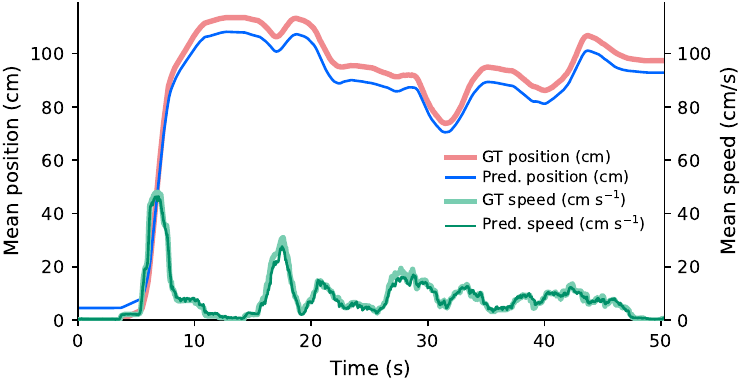}

\caption{\textbf{Mean predicted and ground-truth position and velocity magnitudes over the validation set.}
Values at each time step are averaged across all validation sequences. The predictions closely follow the ground truth over time, showing that the position estimate does not drift and the velocity estimate remains accurate with time.}
    \label{fig:meanplot}
\end{figure}

\begin{table}[t]
    \centering
    \caption{\textbf{Mean absolute error at different target speeds.}
    Target speed is measured using Kobuki wheel odometry and reported in m/s.}
    \label{tab:speed_mae}
    \setlength{\tabcolsep}{5pt}
    \renewcommand{\arraystretch}{1.10}
    \small
    \begin{tabular}{ccccc}
        \toprule
        \textbf{Speed} & $e_x$ & $e_y$ & $e_{v_x}$ & $e_{v_y}$ \\
        (m/s) & (cm) & (cm) & (cm/s) & (cm/s) \\
        \midrule
        0.25 & 3.69 & 3.59 & 0.79 & 0.80 \\
        0.75 & 6.48 & 6.34 & 1.22 & 1.15 \\
        1.00 & 6.65 & 6.37 & 1.35 & 1.28 \\
        \bottomrule
    \end{tabular}
\end{table}

\begin{figure*}[t]
    \centering
    \includegraphics[
        width=0.95\linewidth
    ]{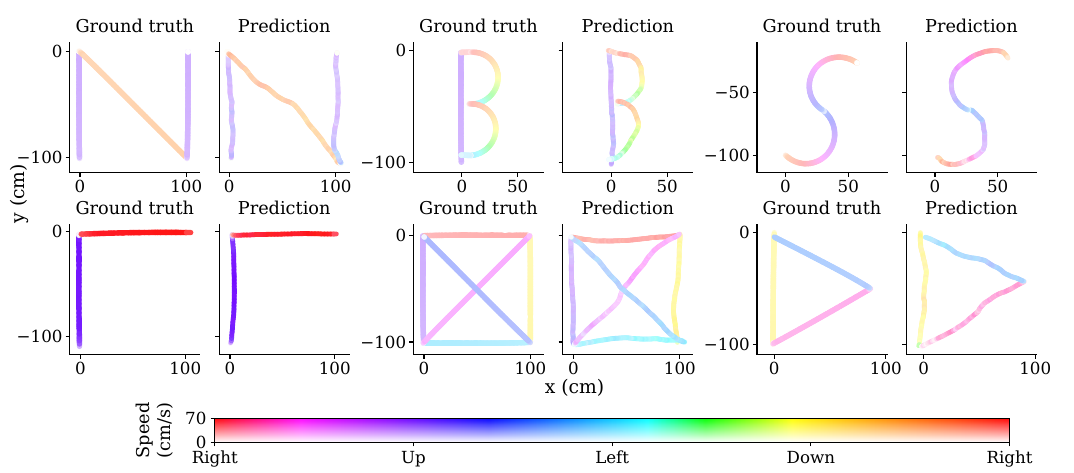}
    \caption{\textbf{Qualitative tracking results on six distinct real-world trajectories.}
For each trajectory, the left panel shows the ground-truth position and color-coded velocity, while the right panel shows filtered position estimates colored by the velocity estimated by our pipeline. The results qualitatively show that the proposed method accurately estimates both position and velocity across diverse real world trajectories.}
    \label{fig:qualitative_results}
\end{figure*}

\begin{figure*}[t]
    \centering
    \includegraphics[width=0.95\textwidth]{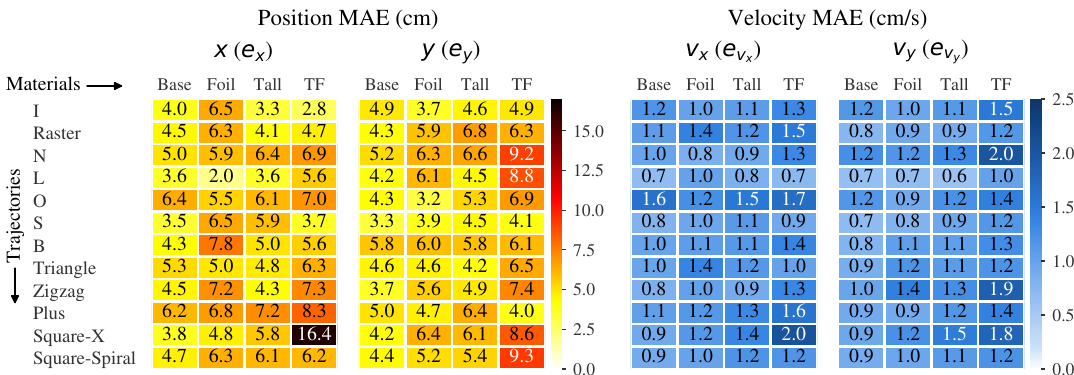}
    \caption{\textbf{Trajectory-wise mean absolute error across different material configurations.}
    Position errors ($e_x$, $e_y$) are reported in cm and velocity errors ($e_{v_x}$, $e_{v_y}$) in cm/s; lighter cells denote lower error. Each cell is averaged over three target speeds and four starting positions, and all errors are reported after applying motion-consistent trajectory refinement.}
    \label{fig:trajectory_material_mae}
\end{figure*}

\begin{table}[t]
    \centering
\caption{\textbf{Ablation study of radar input dimensions and temporal memory.}
To ablate azimuth, Doppler, or range, we average the power and validity mask along that dimension, broadcast the mean back to the original shape, and zero the corresponding coordinate channel. To remove temporal memory, we reset both GRU hidden states before each frame.}
    \label{tab:ablation_study}

    \setlength{\tabcolsep}{2pt}
    \renewcommand{\arraystretch}{1.10}

    \begin{tabular}{lcc}
        \toprule
        \textbf{Configuration}
        & \textbf{Position MAE (cm)}
        & \textbf{Velocity MAE (cm/s)} \\
        \midrule

        \textbf{Full model}
        & \textbf{8.95}
        & \textbf{1.75} \\

        w/o GRU Temporal Memory
        & 11.16
        & 2.10 \\
        
        w/o Azimuth
        & 40.82
        & 3.39 \\

        w/o Doppler
        & 71.22
        & 8.35 \\

        w/o Range
        & 46.94
        & 4.04 \\

        \bottomrule
    \end{tabular}
\end{table}

\begin{figure}[t]
    \centering
    \includegraphics[width=\columnwidth]{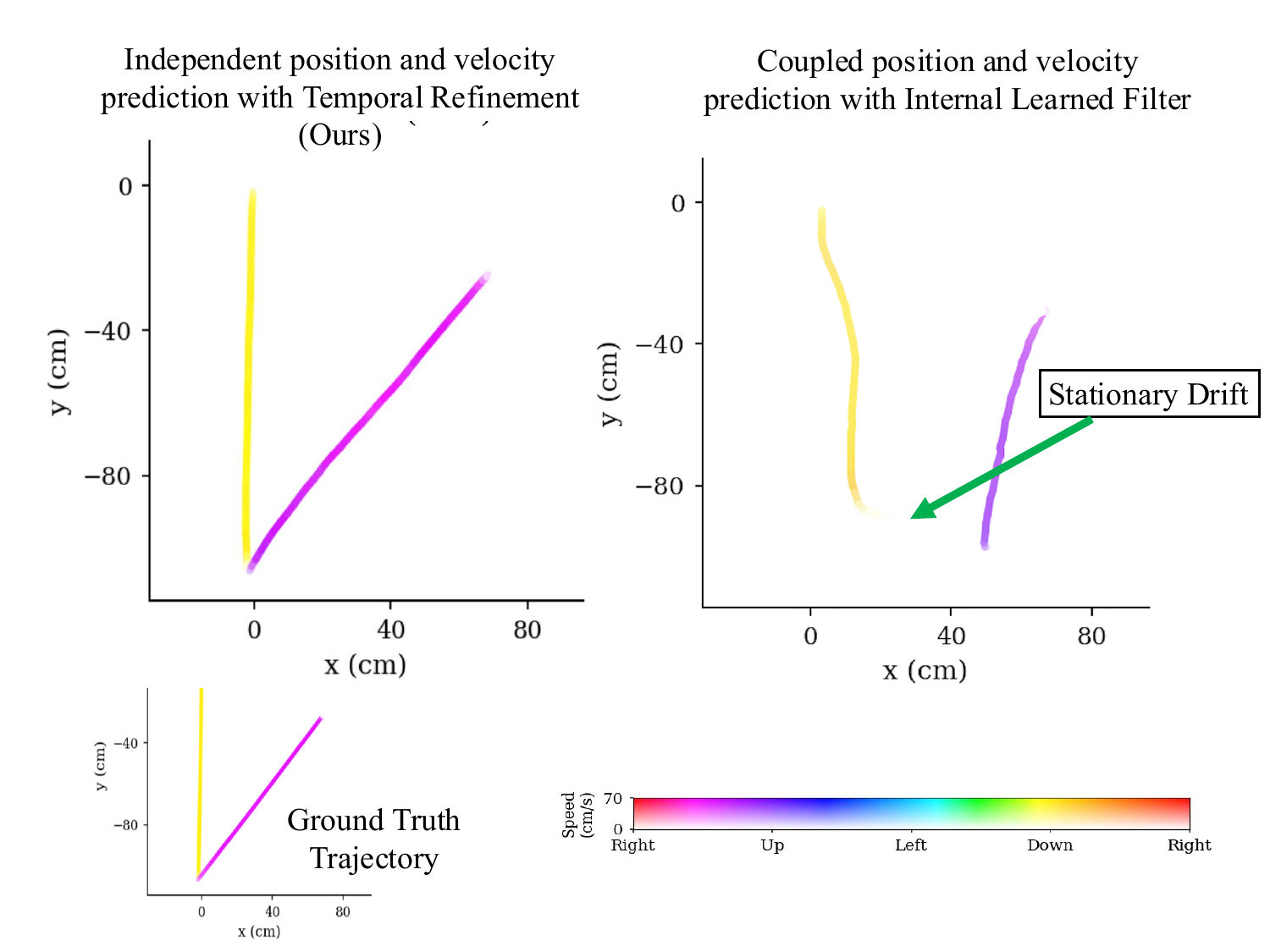}
    \caption{\textbf{Comparison with similar architecture but using an internal learned filter and coupled position-velocity updates.} The coupled model exhibits position drift during stationary intervals, whereas our independent estimation followed by temporal refinement reduces drift. Across the full held-out N-validation set, the coupled model achieves mean position and velocity errors of $21.79$\,cm and $3.27$\,cm/s, compared with $8.95$\,cm and $1.75$\,cm/s for ours.}
    \label{fig:modelablation3}
\end{figure}

\vspace{-1mm}
\section{Results}

\label{sec:results}

\subsection{Localization and Velocity Accuracy}
We train on 528 recordings and evaluate on 48 held-out N-shaped recordings. We report mean Euclidean position and velocity errors.
Over the 48 recordings validation set the network achieves a position error of $15.59$ cm and a velocity error of $2.01$ cm/s. Motion consistent trajectory refinement reduces these errors to $8.95$ cm and $1.75$ cm/s, respectively, improving position accuracy by $42.6\%$ and velocity accuracy by $13.1\%$. We also report component-wise position and velocity MAE across all 12 trajectories dataset and four target configurations—Base, Foil, Tall, and TF—averaged over three target speeds and four starting positions, as shown in Fig.~\ref{fig:trajectory_material_mae}. Qualitative results in Fig.~\ref{fig:qualitative_results} further illustrate the agreement between predicted and ground-truth trajectories, with the color-coded velocity profiles showing consistent velocity tracking across diverse motion patterns. Fig.~\ref{fig:meanplot} further shows the estimator's temporal behavior on the validation set. The predicted position and velocity magnitudes closely follow the corresponding ground-truth profiles throughout the trajectories, with no progressive increase in position error over time. This indicates that the temporal model maintains stable position estimates while preserving the underlying target motion. We also notice that the estimation error increases with target speed, as summarized in Table~\ref{tab:speed_mae}. The lowest position and velocity errors are observed at $0.25$\,m/s, while the errors remain relatively similar between $0.75$ and $1.00$\,m/s, indicating stable tracking performance at higher target speeds.

\subsection{Analysis of Trajectory Refinement}
To directly evaluate whether separating position and velocity estimation before refinement is beneficial, we compare our approach with a closely matched architecture that instead couples position and velocity through an internal learned filter. As shown in Fig.~\ref{fig:modelablation3}, the coupled formulation tends to accumulate position offsets, particularly during stationary intervals. This occurs because accurate velocity alone cannot correct an erroneous absolute position estimate: once the position drifts, a near-correct velocity prediction may preserve the offset rather than recover the true location.
Consistent with this behavior, the largest improvement from our refinement occurs during stationary intervals, when position predictions can fluctuate despite low predicted velocity. Using a ground-truth speed threshold of $1$ cm/s to separate moving and stationary frames for analysis, position error decreases from $16.27$ cm to $8.84$ cm while stationary. During movement, it decreases from $12.31$ cm to $9.49$ cm. Thus, the reported improvement does not depend on discarding stationary frames or evaluating only favorable portions of a trajectory.
These results highlight the complementary roles of position and velocity. Position predictions provide an absolute location estimate, whereas velocity constrains changes between successive locations. Integrating velocity alone cannot correct an existing position offset, while independently using position predictions permits frame-to-frame jitter. Our trajectory refinement balances both sources of information over the recording, suppressing inconsistent fluctuations without relying exclusively on either output.

\begin{figure}[t]
    \centering
    \includegraphics[
        width=\linewidth
    ]{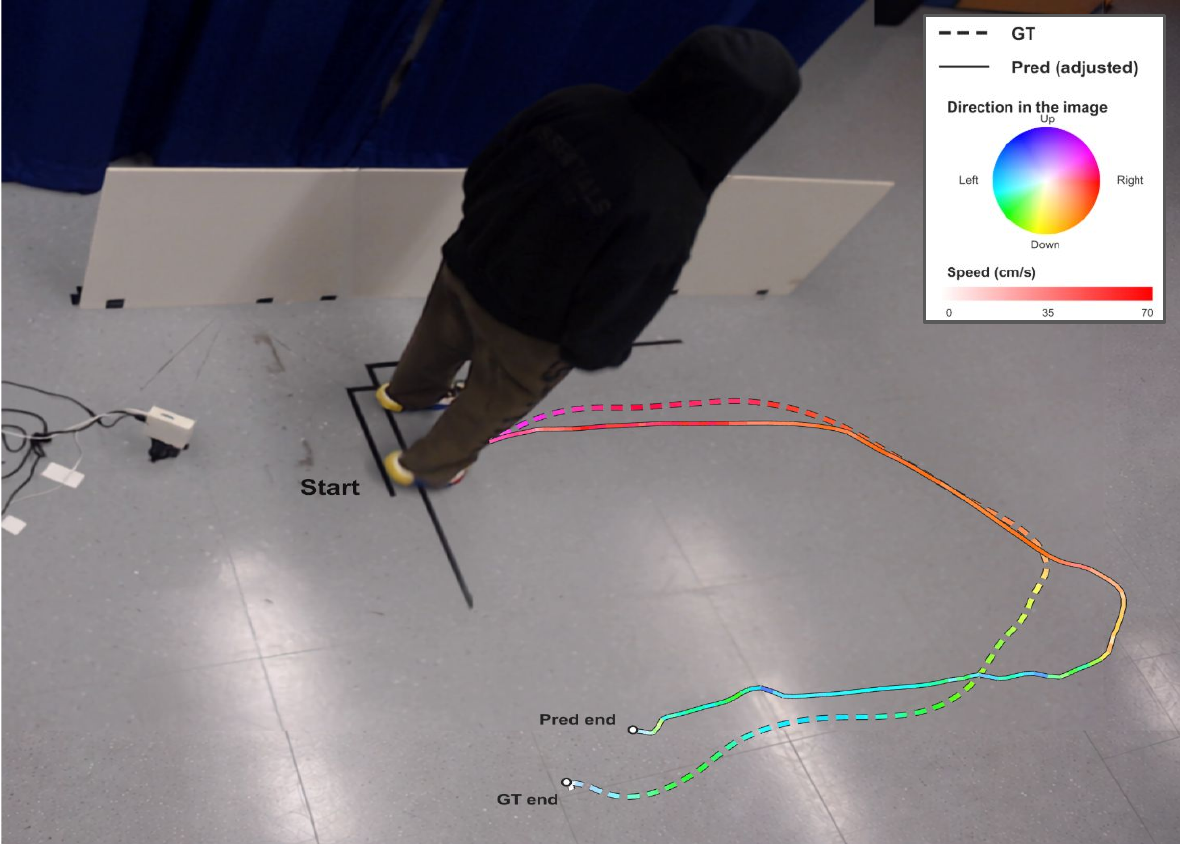}
    \caption{\textbf{Evaluation of generalization across unseen targets and trajectories.} A walking person (training used only the TurtleBot) on a trajectory absent from training; Our model approximately recovers the position and velocity trajectory, demonstrating good out-of-distribution generalization.}
    \label{fig:generalization}
\end{figure}

\subsection{Ablation Study}

Table~\ref{tab:ablation_study} evaluates the contributions of radar input dimensions and temporal memory over all 48 validation recordings. To ablate azimuth, Doppler, or range, we average the preprocessed power and validity-mask inputs along the corresponding dimension, broadcast the mean back to the original tensor shape, and zero its coordinate channel. This removes axis-dependent variation while preserving the input dimensions. Removing Doppler produces the largest degradation, increasing position error from 8.95 to 71.22\,cm and velocity error from 1.75 to 8.35\,cm/s, demonstrating its importance for both localization and motion estimation. Removing azimuth or range increases position error to 40.82 and 46.94\,cm, respectively, indicating substantial dependence on all three radar dimensions. To isolate temporal memory, we reset both GRU hidden states before every frame while retaining their learned weights. Retaining history reduces filtered position error from 11.16 to 8.95\,cm (19.8\%) and velocity error from 2.10 to 1.75\,cm/s (16.7\%). Across speed bins, temporal memory provides larger improvements in velocity estimation at higher speeds, reducing mean error by $0.36$ and $0.87$\,cm/s in the $40$--$60$ and $60$--$80$\,cm/s bins, respectively.

\subsection{Comparison with Related Methods}

We contextualize our results against representative radar-based NLoS localization methods. CornerRadar~\cite{yue2022cornerradar} incorporates scene-derived hint maps and reports a median localization error of $0.263$\,m in settings without a direct propagation path, whereas our method does not require explicit scene geometry at inference. Scheiner et al.~\cite{scheiner2019seeing} exploit Doppler information to estimate target position and motion, reporting a mean localization error of $0.120$\,m. Xu et al.~\cite{xu2024tracking} report a localization error of $0.1092$\,m using range--angle inputs followed by Kalman filtering, and observe degraded localization when range--Doppler information is additionally incorporated. In contrast, our method preserves the full range--azimuth--Doppler representation and jointly estimates position and velocity from sequential radar observations, followed by motion-consistent trajectory refinement. On the held-out N-trajectory validation set, it achieves mean position and velocity errors of $0.0895$\,m and $0.0175$\,m/s, respectively. Under the same experimental setting, our learned model also improves substantially over our geometric baseline, which selects a moving radar peak and projects it assuming a perfectly reflective wall, yielding a mean position error of $0.3988$\,m. These results are intended as contextual comparisons rather than direct rankings, since the baseline methods differ in sensing hardware, scene geometry, input representations, evaluation protocols, reported error statistics, and use of temporal information.

\section{Conclusion}
We presented a learning-based framework for indoor NLoS position-velocity tracking using FMCW radar. Our method preserves the full Range--Azimuth--Doppler (RAD) representation and exploits consecutive radar observations to jointly estimate the position and velocity of a hidden moving target. To train this method, we collected a real-world NLoS radar dataset that covers diverse trajectories, motion speeds, starting positions, and target configurations, with synchronized position and velocity ground truth. Our setting focused on a single relay wall, fixed radar, and one moving target. Future work could investigate more diverse environments and propagation geometries to improve generalization beyond the current experimental setting. We believe that by learning to leverage Doppler information from commodity radars for NLoS tracking, our work opens new possibilities for safe indoor autonomous navigation.


\enlargethispage{1.0mm}
\bibliographystyle{IEEEtran}
\bibliography{references}

\end{document}